%% file: main.tex
\documentclass[11pt]{article} 
\usepackage{acl}

\input{math_commands.tex}

\usepackage{times}
\usepackage{latexsym}

\usepackage[T1]{fontenc}

\usepackage[utf8]{inputenc}

\usepackage{microtype}

\usepackage{inconsolata}

\usepackage{hyperref}
\usepackage{cleveref}
\usepackage{url}
\usepackage{booktabs}
\usepackage{graphicx}
\usepackage{xspace}
\usepackage{array}
\usepackage{multirow}
\usepackage{multicol}
\usepackage{subcaption}

\title{Principled Content Selection to Generate Diverse and Personalized Multi-Document Summaries}

\author{Vishakh Padmakumar$^{1}$\thanks{Work done during summer internship at Adobe}~~~~~Zichao Wang$^{2}$~~~~~David Arbour$^{2}$~~~~~Jennifer Healey$^{2}$ \\  
 $^1$New York University~~~~~$^2$Adobe Research\\
 {\tt\small vishakh@nyu.edu} \\
}

\begin{document}
\input{macros}

\maketitle

\begin{abstract}
While large language models (LLMs) are increasingly capable of handling longer contexts, recent work has demonstrated that they exhibit the \emph{"lost in the middle"} phenomenon \cite{liu2024lost} of unevenly attending to different parts of the provided context. This hinders their ability to cover diverse source material in multi-document summarization, as noted in the \diversumm benchmark \cite{huang2024embrace}. In this work, we contend that principled content selection is a simple way to increase source coverage on this task. As opposed to prompting an LLM to perform the summarization in a single step, we explicitly divide the task into three steps---(1) reducing document collections to atomic key points, (2) using determinantal point processes (DPP) to perform select key points that prioritize diverse content, and (3) rewriting to the final summary. By combining prompting steps, for extraction and rewriting, with principled techniques, for content selection, we consistently improve source coverage on the \diversumm benchmark across various LLMs. Finally, we also show that by incorporating relevance to a provided user intent into the DPP kernel, we can generate \emph{personalized} summaries that cover \emph{relevant} source information while retaining coverage. 
\end{abstract}

\input{intro}
\input{formulation}
\input{experiments}

\input{results}

\input{related}
\input{conclusion}

\section*{Acknowledgments}
We would like to thank Nishant Balepur, Yoonjoo Lee, Dayeon Ki, Dang Nguyen, Paiheng Xu, Hyunji Lee, Hung-Ting Chen, Nitish Joshi, Tong Sun, and Vishy Swaminathan for their feedback at various stages of the project. 
This work was completed while Vishakh was an intern at Adobe. At NYU, Vishakh is supported by the National Science Foundation under Grant No. IIS-2340345 and Grant No. 1922658. 

\bibliography{iclr2024_conference}

\appendix
\input{appendix}

\end{document}

%% file: math_commands.tex
\usepackage{amsmath,amsfonts,bm}

\def\eqref#1{equation~\ref{#1}}

\def\1{\bm{1}}

\DeclareMathAlphabet{\mathsfit}{\encodingdefault}{\sfdefault}{m}{sl}
\SetMathAlphabet{\mathsfit}{bold}{\encodingdefault}{\sfdefault}{bx}{n}



%% file: macros.tex
\newcommand\todo[1]{\textcolor{red}{[TODO: #1]}}
\newcommand\vis[1]{\textcolor{violet}{[VP: #1]}}
\newcommand\jw[1]{\textcolor{brown}{[JW: #1]}}
\newcommand\jh[1]{\textcolor{purple}{[JH: #1]}}
\newcommand\da[1]{\textcolor{blue}{[DA: #1]}}

\newcommand{\PreserveBackslash}[1]{\let\temp=\\#1\let\\=\temp}
\newcolumntype{C}[1]{>{\PreserveBackslash\centering}p{#1}}
\newcolumntype{R}[1]{>{\PreserveBackslash\raggedleft}p{#1}}
\newcolumntype{L}[1]{>{\PreserveBackslash\raggedright}p{#1}}

\definecolor{red}{RGB}{255, 0, 0}
\definecolor{blue}{RGB}{135, 206, 250}
\definecolor{green}{RGB}{205, 255, 204}

\newcommand{\eg}{e.g.,\xspace}
\newcommand{\ie}{i.e.\xspace}
\newcommand{\diversumm}{\textsc{Diversesumm}\xspace}
\newcommand{\ourmethod}{\emph{LLM + DPP}\xspace}
\newcommand{\llmsimple}{\emph{Naive LLM}\xspace}
\newcommand{\llmall}{\emph{All KPs}\xspace}
\newcommand{\llmllm}{\emph{LLM-Selected KPs}\xspace}

\newcommand{\redcell}{\cellcolor[HTML]{ffad99}}
\newcommand{\greencell}{\cellcolor[HTML]{99ff99}}
\newcommand{\lightredcell}{\cellcolor[HTML]{ffd6cc}}
\newcommand{\lightgreencell}{\cellcolor[HTML]{ccffcc}}

%% file: intro.tex
\section{Introduction}

\begin{figure*}[ht!]
    \centering
    \includegraphics[width=\linewidth]{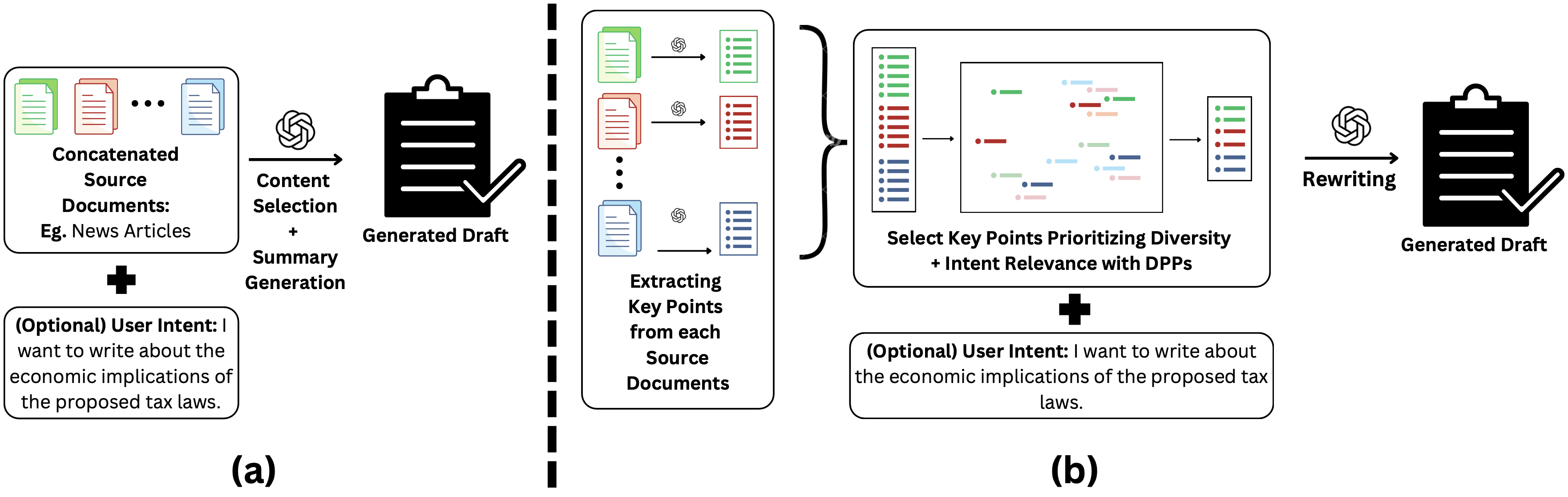}
    \caption{Overview of the MDDS task (\Cref{sec:formulation}), which aims to generate a summary from a set of source documents with an optional user intent. Compared to (a) prompting an LLM to perform MDDS in a single step (\llmsimple) and other baselines, (b) our method (\ourmethod) first extracts atomic key points from each document, then explicitly selects content using DPPs to ensure diversity and relevance before rewriting them into a summary (\Cref{sec:method}). \ourmethod improves source coverage and produces summaries more aligned with user intent (\Cref{sec:results}).}
    \label{fig:fig1}
\end{figure*}

Recent advances in language modeling have enabled contemporary models to handle very long contexts \citep{reid2024gemini, anthropic-claude}, spurring new evaluations of their capabilities in these settings \citep{tay2021long, pang2022quality, shaham2022scrolls, kamradt2023needle, karpinska2024one}. As it becomes possible to process these longer inputs, \citet{zheng2024lmsys} observe that a common use case of LLMs involves the summarization of dense information from collections of documents.    
A key challenge in providing reliable output for the users in this setting is ensuring high coverage of the source material when multiple documents present diverse viewpoints on the same issue---a problem formalized by the \diversumm benchmark \citep{huang2024embrace} as Multi-Document Diversity Summarization (MDDS). 

While contemporary models are highly capable, their attention mechanisms tend to prioritize content at the start and end of the context \cite{liu2024lost}. This bias is particularly problematic for MDDS, where key details may be 
spread across multiple documents. As a result, even state-of-the-art LLMs like GPT-4 struggle when prompted to complete the MDDS task \cite{huang2024embrace} despite performing well on single-document summarization \cite{goyal2022news}, where clear introductions and conclusions provide natural focal points.
Furthermore, deploying LLMs in public-facing interfaces highlights another important facet of the MDDS problem---ensuring reliable coverage of all \emph{relevant} information in a collection of documents to user intents, essentially an instance of \emph{Query-focused} summarization \citep{daume2006bayesian, vig2022exploring}. There exists an open question to investigate how the attention biases of LLMs interact with information relevance to user intents when generating summaries.

Our research question is: How does content selection impact the source coverage of LLMs in MDDS? (\Cref{sec:method}). We observe that prompting an LLM for the task involves implicitly selecting relevant content and generation into a coherent summary in a single step. Instead, we decouple this single prompting step into principled content selection to prioritize diversity, defending against the aforementioned attention bias, followed by a rewriting step to produce a coherent, high-coverage summary (\Cref{fig:fig1}).

In order to select content, we draw inspiration from recent work
which shows that LLMs reliably break down individual documents into atomic claims or key points \cite{fables24kim, padmakumar2023does, krishna2023longeval}. 
After extracting key points from each source document, 
we use \emph{determinantal point processes} (DPPs) \cite{kulesza2012determinantal}
to select the subset of key points used to generate the summary.  
DPPs are a statistical model that are used to select subsets of items prioritizing diversity.\footnote{We detail related work that uses DPPs in recommender systems (\Cref{sec:related-dpp-apps}) and as well as previous approaches to single document summarization (\Cref{sec:related-dpps-summarization}) 
} Finally, we rewrite the selected key points into the desired output by prompting an LLM. 

We show that using DPPs for diverse content selection consistently improves coverage on the \diversumm benchmark, compared to both a naive prompting baseline and a multi-step LLM-prompt pipeline, robustly across multiple LLMs---GPT-3.5, GPT-4o, Claude-3-Sonnet, and Llama 3.1 (\Cref{sec:multi-doc-summ-results}). Content selection via DPPs can also be tuned to incorporate a relevance matrix generating summaries that are better aligned with user intents (\Cref{sec:relevance-relevance}). 
As LLMs are increasingly deployed in sequential, agentic pipelines for complex tasks, our findings show the value of complementing LLM prompting steps—such as extracting and rewriting key points—with principled techniques like DPPs for content selection, where appropriate, to achieve stronger performance.

%% file: formulation.tex
\section{Problem Formulation}
\label{sec:formulation}

\paragraph{Multi-document diversity summarization (MDDS)}

The MDDS task, as formulated by \citet{huang2024embrace}, focuses on generating a summary $s$ from a set of articles, $D = \{d_{1 \dots k}\}$, covering the same news story.  
Each set $D$ is paired with a set of questions $Q = \{q_{1}, \dots, q_{m}\}$, which contain diverse answers drawn from multiple source documents. The objective is to model $p(s|D)$ such that the summary $s$ is faithful to the source content and achieves high coverage, as measured by correctly answering a large number of questions $q_i \in Q$ based on the summary $s$.

\paragraph{Query-focused Multi-document diversity summarization}  
Building on the MDDS framework, we also explore a variation known as \emph{query-focused} summarization \citep{daume2006bayesian}. In this task, the input consists of the set of articles $D$ and a user-specified query $q_{\text{user}}$. The goal is to model $p(s | D, q_{\text{user}})$, where the summary $s$ has high coverage of content \emph{relevant} to the $q_{\text{user}}$. Relevance is determined using a scoring function $f_{\text{rel}}(q_i | q_{\text{user}})$, which identifies the subset of relevant questions $Q_{\text{user}} \subset Q$. We evaluate the summary based on coverage of relevant questions $q_i \in Q_{\text{user}}$. 

\section{Constructing Documents With Principled Key Point Selection}
\label{sec:method}

A typical LLM pipeline for summarizing long contexts involves either concatenating multiple source documents and performing summarization via a single zero-shot prompt \cite{huang2024embrace}, or hierarchically summarizing the collection, using prompting to process individual documents \cite{chang2024booookscore}. 
We hypothesize that LLMs might not be well suited to perform the content selection aspect of summarization. 
To test this, we design a three-step pipeline (\Cref{fig:fig1}) that 
constructs a summary by extracting atomic key points from each document (\Cref{sec:extraction}), selects the key points to be included in the summary in a principled manner, prioritizing diversity of content (\Cref{sec:kp_div_selection}) as well as relevance to a user intent (\Cref{sec:kp_rel_selection}) and then rewrites the selected key points into the summary (\Cref{sec:rewriting}). We then evaluate the coverage of summaries generated with our method (\Cref{sec:evaluation_method}) to various baselines (\Cref{sec:models}).

\subsection{Key Point Extraction}
\label{sec:extraction}
Given a set of documents $D = \{d_{1 \dots k}\}$, we use an LLM to decompose each document 
$d_i$ into a set of key points $K_i = \{k_{i,1}, k_{i,2} \dots k_{i,n}\}$ that represent distinct pieces of information within the text. Prior work has demonstrated that LLMs reliably break down individual documents into atomic claims or key points via a zero-shot prompt for various applications \citep{fables24kim, padmakumar2023does, krishna2023longeval}. We aim to generate a summary $s$ that allows for high coverage of $Q$ associated with $D$. Each extracted key point captures an atomic claim or distinct piece of information, so we hypothesize that selecting diverse key points would lead to better coverage of $Q$.

\subsection{Principled Key Point Selection}
\label{sec:selection}
Given the set of all key points from all documents, $ K = \bigcup_i K_i $, the next step involves selecting a subset of key points, $K_{sel}$, prioritizing coverage of source material for MDDS, additionally incorporating relevance for \emph{query-focused} summarization. 

\subsubsection{Background on DPPs}
\label{sec:dpp_background}
Determinantal Point Processes (DPPs) model the probability of selecting subsets from a set of items emphasizing diversity among the chosen elements \cite{kulesza2012determinantal}. DPPs construct a kernel matrix $L$ using a similarity function between pairs of items. The kernel matrix may also be weighted by a diagonal matrix that scores the absolute quality or a task-specific property such as the relevance of the items \cite{kulesza2012determinantal}.  Inference from DPPs is formulated as a combinatorial optimization problem, where the goal is to find the subset of items with the highest likelihood under the kernel $L$. This can be efficiently approximated using greedy algorithms \citet{chen2018fast}.  
Our work uses DPP inference out of the box, noting that this allows the number of selected items to vary according to the similarity of items in the kernel matrix rather than a pre-specified number of distinct items. 
We provide more extensive coverage of prior work connecting DPPs with NLP tasks in \Cref{sec:related-dpps-summarization}. 

\subsubsection{Selecting Key Points Prioritizing Diversity}
\label{sec:kp_div_selection}
To achieve high source coverage in the MDDS task, we use a DPP to select a subset of key points from $ K = \bigcup_i K_i $ that prioritizes diversity. Each key point $ k_{ij} $ is first embedded into a high-dimensional vector $ v_{ij} $ via a transformer-based encoder. These embeddings are then used to construct a kernel matrix $ L $, where each entry $ L_{(i_1,j_1),(i_2,j_2)} $ represents the similarity between pairs of key points, computed through a kernel function $ f_k(v_{i_1j_1}, v_{i_2j_2}) $. We then run DPP-inference on L to obtain the selected key points, $K_{sel}$ as detailed in \Cref{sec:dpp_background}.

\subsubsection{Selecting Relevant Key Points Prioritizing Diversity}
\label{sec:kp_rel_selection}
In the \emph{query-focused} MDDS task, we incorporate relevance to 
$q_{user}$
into the key point selection objective, using a modified DPP approach. After embedding each key point $k_{ij}$ into a vector $v_{ij}$, we construct the similarity matrix $L$ as above.
We then create the relevance vector $R$, where each entry $R_i$ represents the relevance score of $k_i \in K$ calculated as $f_{\text{rel}}(k_i|q_{\text{user}})$.\footnote{We note here that the dimensionality of $R$ is equal to the total number of key points across \emph{all} source documents, the same as that of $L$.} The relevance-weighted matrix to $L' = R L R^T$ thus balances both key point similarity and relevance to $q_{user}$. where each entry in $L'_{(i_1,j_1),(i_2,j_2)} = f_{rel}(v_{i_1j_1} | q_{user}) \times f_k(v_{i_1j_1}, v_{i_2j_2}) \times f_{rel}(v_{i_2j_2}|q_{user})$. DPP inference is then applied to $L'$ (\Cref{sec:dpp_background}), selecting a diverse yet query-relevant subset 
$K_{sel}$.

\subsection{Rewriting}
\label{sec:rewriting}
The final step involves synthesizing the selected key points into a coherent summary 
$s$. We use an LLM to rewrite the chosen subset, ensuring that the output is coherent and well-structured. 

%% file: experiments.tex
\section{Experimental Setup}
\label{sec:expt}

\subsection{Datasets}
\label{sec:data}

\subsubsection{DiverseSumm Benchmark}
\label{sec:diversumm_org}
The \textsc{DiverseSumm} benchmark consists of 245 examples, each of which is a set 10 articles covering different aspects of the same news event. Each example is accompanied by $1$ to $10$ questions, with each question linked to a set of articles that provide answers. These articles offer diverse perspectives on the questions, and the objective is to produce a summary that captures the range of perspectives. 

\subsubsection{Augmenting DiverseSumm with more questions}
\label{sec:diversumm_aug}
We observe that $78.3\%$ 
of news stories in the original dataset have $3$ or fewer associated questions. Thus not all articles are associated with questions in each example. To better evaluate the coverage of individual articles by the different methods, 
we use GPT-4o to generate $10$ additional questions per article for each news story. This results in a synthetically augmented version of \textsc{DiverseSumm} with 100 questions per news story, sourced from the different articles.\footnote{To verify the quality of the augmented questions, we conduct a human annotation in \Cref{sec:diversumm_aug_validation}.} The prompt to obtain these questions is provided in \Cref{sec:prompt_additional_questions}.  
Unlike the original dataset, we do not expect these questions to have coverage across multiple articles, but this helps improve the statistical power of our comparison across methods. We report results on both the original, as well as augmented versions of the \textsc{DiverseSumm} dataset.

\subsubsection{Augmenting DiverseSumm with synthetic user intents}
\label{sec:diversumm_intent}
Finally, to adapt \textsc{DiverseSumm} for a \emph{query-focused} multi-document summarization task, we synthetically generate user intents to accompany each news story.  These user intents reflect varied information needs, making certain perspectives from the source articles more or less relevant based on the intent. We prompt an LLM, again GPT-4o, to produce $5$ distinct user intents for each news story given the concatenated set of $10$ articles.\footnote{We detail the method in which we filter generated user intents for quality as well as identify the set of relevant questions for evaluation in \Cref{sec:diversumm_intent_filtering}. We also conduct a human annotation to verify the quality of the intents in \Cref{sec:diversumm_intent_validation}.} The prompt details for generating these user intents are provided in \Cref{sec:prompt_additional_questions}. 

\subsection{Evaluation}
\label{sec:evaluation_method}

\paragraph{Automatic Evaluation of Source Coverage}

To evaluate the coverage of generated summaries, we measure how many questions $q_i \in Q$ can be correctly answered based on the summary $s$. We evaluate if a question is answered using an LLM-as-judge evaluation with GPT-4o to (a) check whether a given question $q_i$ is answerable from $s$, and (b) verify whether the answer from $s$ aligns with the content in the corresponding article $d_j \in D$. A question $q_i$ is \emph{covered} by $s$ if $q_i$ is answerable from $s$ and if the answer for $q_i$ obtained from $s$ matches the answer from $d_j$. We report the average coverage of examples from \diversumm and \diversumm Augmented (\Cref{sec:data}). Prior work has demonstrated the effectiveness of evaluation of question-answering tasks by prompting an LLM \cite{li2024pedants, balepur2024reverse}. We select the prompt format per the recommendations of \citet{huang2024embrace} to evaluate the coverage of each question individually from the summary and the faithfulness of the answer to the original article, each via binary answers from an LLM. We provide the prompt used in \Cref{sec:prompt_eval_answerability}.

\paragraph{Correlation with Human Judgments}
To validate the reliability of our automatic evaluation, we cross-check a random sample of LLM-as-judge outputs from GPT-4o against human annotations collected via Amazon Mechanical Turk. We sample $100$ outputs, equally split between cases where $q_i$ is answerable from $s$ and cases where it is not, obtaining $3$ human annotations for each. The agreement between the LLM-as-judge and human annotations is $86.4\%$ for answerability and $95.3\%$ for correctness, demonstrating the robustness and reliability of the automatic evaluation method.\footnote{We note that this agreement matches is in line with the reported performance of GPT-4 in an LLM-as-judge setting in MT-Bench \cite{zheng2023judging}. However, we acknowledge the limitations of LLM-as-judge evaluation in \Cref{sec:limitations}.}

\subsection{Models Used}
\label{sec:models}

We perform experiments using four LLMs: GPT-3.5, GPT-4o \cite{openai2024gpt4}, Claude-3-Sonnet \cite{anthropic-sonnet}, and LLaMA 3.1 70B \cite{dubey2024llama}.
\paragraph{Our Method (\ourmethod)} We first perform key point extraction from each article using the respective LLM (\Cref{sec:extraction}) with the prompt detailed in \Cref{sec:prompt_extract_kp}. We then select the key points to be included in the summary (\Cref{sec:selection}) using the DPPy library \cite{gautier2019dppy}.\footnote{We perform exact sampling via the spectral method, the default inference technique via DPPy} We create the Gaussian kernel matrix $L$, using BertScore \cite{zhang2020BERTScore} with \texttt{Deberta-V3} embeddings \cite{hedebertav3} as the similarity function between pairs of key points---we ablate aspects of the DPP kernel in \Cref{tab:results_kernel_variants}. Additionally, we score the relevance of different key points to $q_{user}$ using an instruction-tuned retrieval model, \texttt{intfloat/e5-mistral-7b-instruct} \cite{wang2023improving} model due to its strong performance on the  \href{https://huggingface.co/spaces/mteb/leaderboard}{MTEB leaderboard}. The selected key points are then rewritten into the final summary using the same LLM (\Cref{sec:rewriting}) with the prompt detailed in \Cref{sec:prompt_rewriting}. 

\paragraph{Baselines} (1) \llmsimple - A simple baseline where we prompt the LLM to generate the summary from the concatenated set of articles, performing content selection and text generation in one step. The prompt for \llmsimple is provided in \Cref{sec:prompt_for_llmsimple}, (2) \llmall - To ablate the effect of our content selection methods, we compare to a baseline where we prompt the LLM to generate the summary from the set of all key points extracted from the articles. This uses the same prompt as \ourmethod for rewriting (\Cref{sec:prompt_rewriting}), just without the selection step, and (3) \llmllm - Finally, to demonstrate the effectiveness of the DPP-based key point selection method over an entirely LLM-prompting pipeline, we compare to a baseline that performs key point selection with an LLM (GPT-4o) before rewriting. The prompt used for LLM-based key point selection is provided in \Cref{sec:prompt_for_llmllm}. This uses the same prompt as \ourmethod for rewriting (\Cref{sec:prompt_rewriting}).

%% file: results.tex
\section{Results}
\label{sec:results}

\begin{table*}[]
\centering
\small
\begin{tabular}{@{}rcccccccc@{}}
\toprule
      & \multicolumn{4}{c}{\diversumm}& \multicolumn{4}{c}{\diversumm Augmented}          \\ \midrule
\multicolumn{1}{c}{}      & \multicolumn{1}{c}{GPT 3.5} & \multicolumn{1}{c}{GPT 4o} & \multicolumn{1}{c}{Claude} & \multicolumn{1}{c}{Llama}  & \multicolumn{1}{c}{GPT 3.5} & \multicolumn{1}{c}{GPT 4o} & \multicolumn{1}{c}{Claude} & \multicolumn{1}{c}{Llama}  \\ \midrule
\multicolumn{1}{r}{\llmsimple} & \multicolumn{1}{c}{0.3324}  & \multicolumn{1}{c}{0.5516} & \multicolumn{1}{c}{0.4776} & \multicolumn{1}{c}{0.2427} & \multicolumn{1}{c}{0.2667}  & \multicolumn{1}{c}{0.4807} & \multicolumn{1}{c}{0.4248} & \multicolumn{1}{c}{0.2187} \\ \midrule
\llmall & 0.3472 & 0.5443& 0.5683& 0.3458& 0.2573 & 0.4620 & 0.4114     & 0.2368     \\
\llmllm & 0.4370 & 0.5747& 0.5369& 0.3376& 0.3849 & 0.5409& 0.5142& 0.3087\\
\midrule 
\ourmethod & 0.4706 & 0.5805& 0.5923& 0.3653& 0.3845 & 0.5535& 0.5469& 0.3227\\
\bottomrule
\end{tabular}
\caption{Source coverage evaluation (\Cref{sec:evaluation_method}) on \diversumm (\Cref{sec:diversumm_org}) and \diversumm Augmented (\Cref{sec:diversumm_aug}). We report coverage of the source material as the fraction of questions correctly answered from the generated summaries (\Cref{sec:evaluation_method}) from 4 different LLMs---GPT3.5, GPT-4o, Claude-3-Sonnet and Llama-3.1, and compare the performance of our method, \ourmethod, with three relevant baselines (\Cref{sec:models}). Selecting key points to prioritize diversity via DPPs (\Cref{sec:selection}) results in better source coverage for all 4 LLMs.}
\label{tab:results_cov}
\end{table*}

\begin{table*}[]
\centering
\small
\begin{tabular}{@{}rcccccc@{}}
\toprule
      & \multicolumn{3}{c}{\diversumm}& \multicolumn{3}{c}{\diversumm Augmented}          \\ \midrule
\multicolumn{1}{c}{}      & \multicolumn{1}{c}{GPT 3.5} & \multicolumn{1}{c}{GPT 4o} & \multicolumn{1}{c}{Claude}  & \multicolumn{1}{c}{GPT 3.5} & \multicolumn{1}{c}{GPT 4o} & \multicolumn{1}{c}{Claude} \\ \midrule
\llmllm & 0.4370 & 0.5747& 0.5369& 0.3849 & 0.5409& 0.5142\\\midrule
\ourmethod (Gaussian Kernel, $\sigma = 0.1$)  & 0.4494 & 0.6145& 0.6347& 0.3728 & 0.6145& 0.6037 \\
\ourmethod (Gaussian Kernel, $\sigma = 1$) & 0.4706 & 0.5805& 0.5923& 0.3845 & 0.5535& 0.5469\\
\ourmethod (Gaussian Kernel, $\sigma = 10$) & 0.4342 & 0.5906& 0.5198 & 0.3752 & 0.5258& 0.4699 \\
\ourmethod (Linear Kernel) & 0.4653 & 0.5893& 0.5863 & 0.3674 & 0.5518& 0.5450 \\ 
\bottomrule
\end{tabular}
\caption{We report 4 ablations of the DPP kernel used for keypoint selection (\Cref{sec:selection}) for our method, \ourmethod. We evaluate 3 LLMs on 4 different kernels for source coverage (\Cref{sec:evaluation_method}).}
\label{tab:results_kernel_variants}
\end{table*}

\subsection{Evaluating Source Coverage in Multi-document Summarization}
\label{sec:multi-doc-summ-results}

\paragraph{Content selection with DPPs results in better source coverage}
From \Cref{tab:results_cov}, we observe that \ourmethod consistently achieves the highest source document coverage across all evaluated LLMs, outperforming all baselines on both the \diversumm and \diversumm Augmented datasets. The baselines that explicitly select key points (\llmall and \llmllm) generally outperform the naive approach of concatenating articles and prompting the LLM for a summary (\llmsimple) for all LLMs. Additionally, the consistent improvement of \llmllm and \ourmethod over \llmall indicates that simply reducing context length by extracting all key points is insufficient, explicit key point selection is important in order to obtain better coverage.

\paragraph{Encoded model representations of key points provide useful signal for key point selection}
From \Cref{tab:results_kernel_variants}, we also observe that \ourmethod, using variants of the DPP-kernel applied to high-dimensional encoder embeddings, outperforms \llmllm, which performs on explicit key point selection in the text space through LLM prompting. This finding shows the value of using principled techniques, such as diversity-aware key point selection (\Cref{sec:selection}), to perform individual steps in a pipeline instead of performing every step via an LLM prompt. 

\begin{figure*}[h]
    \centering
    \includegraphics[width=\linewidth]{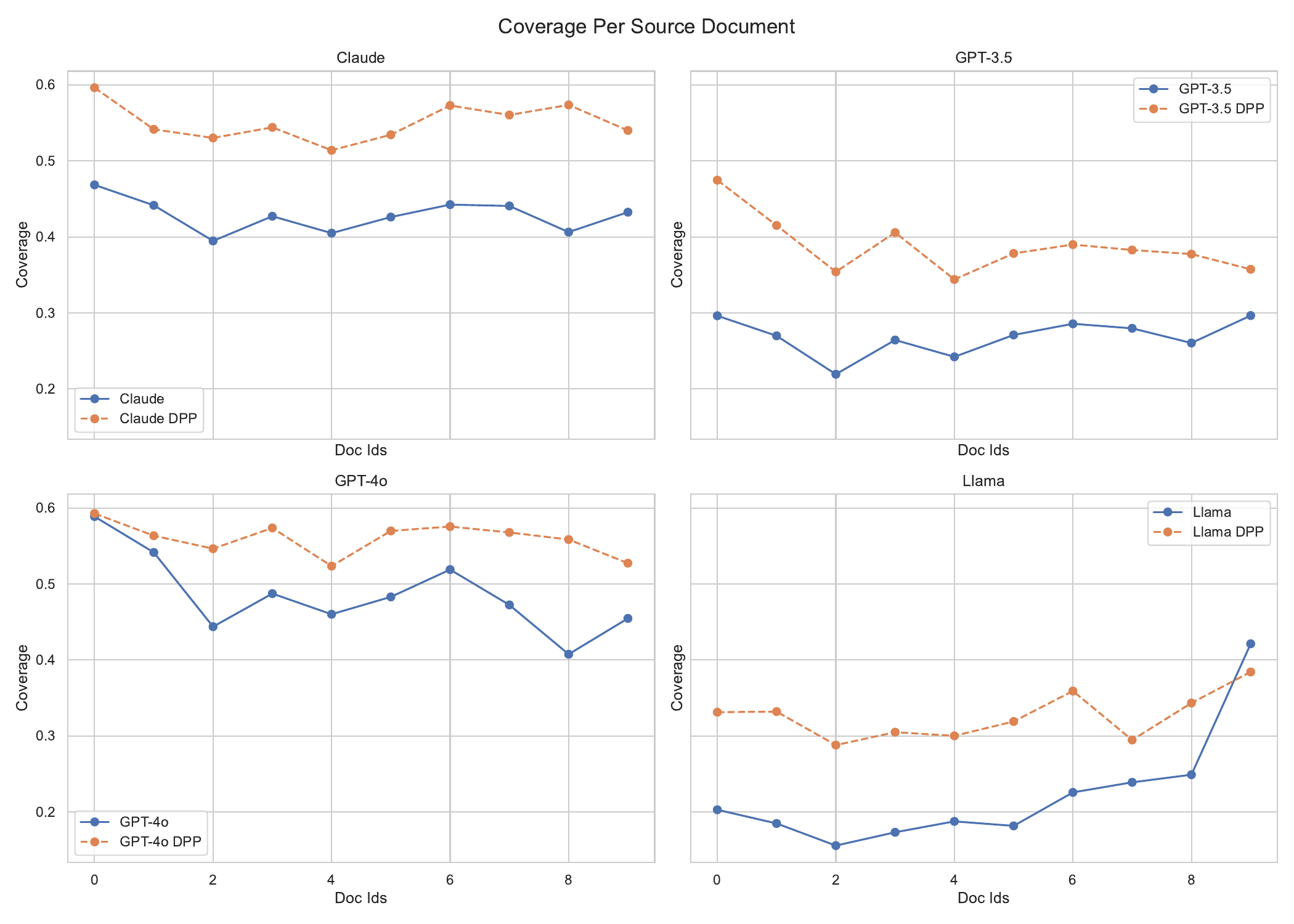}
    \caption{Studying the 'lost-in-the-middle' phenomenon by plotting coverage of different source articles by index with \llmsimple and \ourmethod. While \llmsimple exhibits biases to better cover the articles at the \emph{start} (GPT-4o, GPT-3.5) or \emph{end} (Llama) of the context, \ourmethod has higher and more uniform coverage of all source documents---mitigating these biases. }
    \label{fig:document_wise_coverage}
\end{figure*}

\paragraph{While LLMs selecting content have uneven coverage, key point selection is more uniform} Prior work \cite{liu2024lost} has shown that LLMs have systematic biases in how well they attend to context, better answering questions when relevant information appears at the start or end of the context. \citet{huang2024embrace} also observe similar 'lost-in-the-middle' biases on the multi-document summarization task. To study this, we plot the coverage of the generated summaries from \ourmethod and \llmsimple per article on \diversumm Augmented in \Cref{fig:document_wise_coverage}.\footnote{We selected the augmented version of \diversumm since the synthetic question generation ensures that each article has at least $10$ associated questions. This ensures we have statistical power on our results.} We observe that the \llmsimple approach exhibits systematic positional biases. Llama 3.1 has better coverage of documents at the end of the context, an \emph{end} bias. Similarly, GPT-4o has a \emph{start} bias, and GPT-3.5 and Claude exhibit mild biases to not sufficiently cover documents in the middle. \ourmethod improves coverage on all documents, particularly alleviating the positional biases on Llama 3.1 and GPT-4o, highlighting the efficacy of key point selection in the multi-document summarization task. 

\paragraph{Key points selected in \ourmethod better covers the source documents than \llmllm}
\begin{figure}[h]
    \centering
    \includegraphics[width=\linewidth]{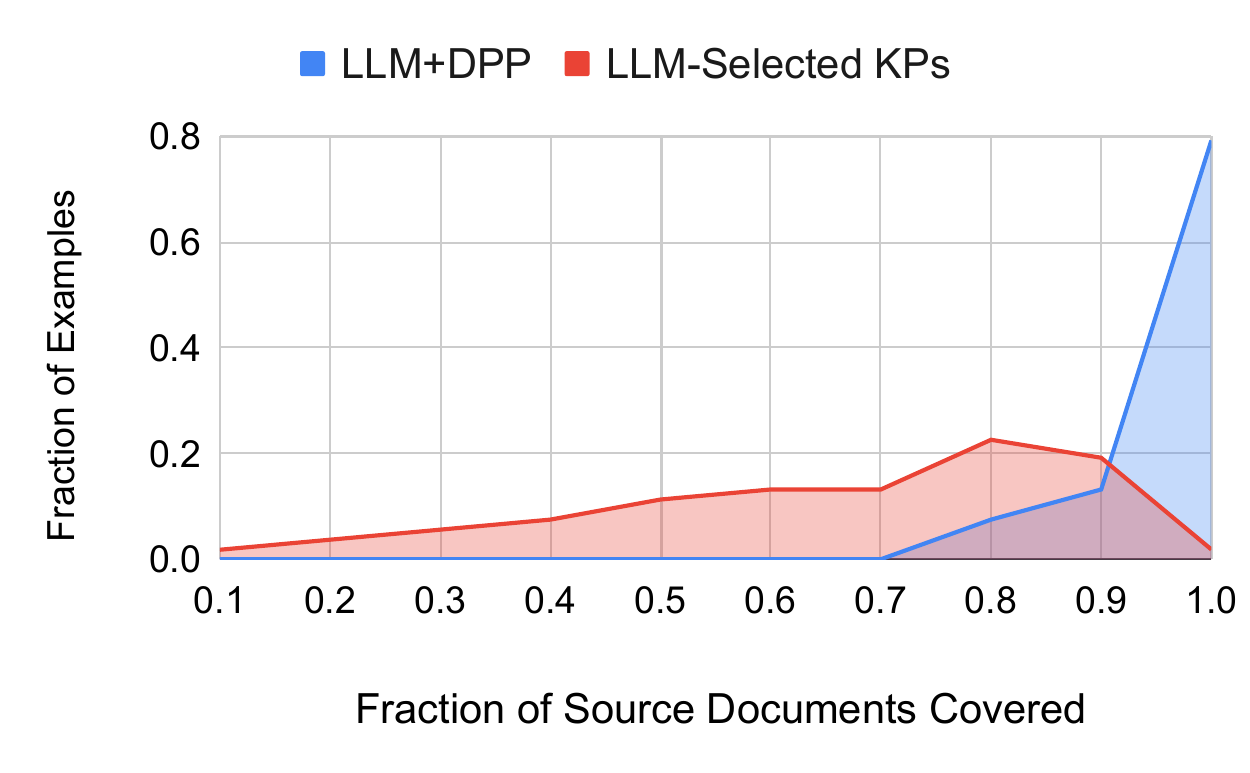}
    \caption{Distribution of source documents covered by key points when selected with \ourmethod and \llmllm. \ourmethod exhibits consistently higher coverage of source documents.}
    \label{fig:source_distribution}
\end{figure}
To investigate the improved coverage of \ourmethod over \llmllm, we plot the distribution of the fraction of source documents covered in the selected subsets of key points in \Cref{fig:source_distribution}. While \llmllm has a much higher variance of documents covered, \ourmethod consistently achieves high coverage of the diverse source documents.\footnote{We perform tests for significance in \Cref{sec:significance-tests}.} 

\paragraph{DPP-based key point selection improves coverage without increasing summary length} To investigate whether the improved source coverage achieved by \ourmethod stems from better content selection rather than simply generating longer summaries---a potential confounder---%
we compare the average summary lengths across \ourmethod, \llmsimple, and \llmllm for each of the four LLMs analyzed (\Cref{tab:lengths}). We calculate the statistical significance of the differences in mean lengths using a two-tailed t-test. We observe no significant differences in average summary lengths for GPT-4o, GPT-3.5, and Llama indicating that the higher source coverage reported in \Cref{tab:results_cov} is not attributable to longer summaries in these.\footnote{For Claude, the differences in summary lengths across the various methods, at odds with the other LLMs, potentially stems from differences in model training---we note that Claude was specifically tuned for long contexts \cite{anthropic-claude}. We believe that this differing behavior when interacting with different inputs for the rewriting step presents a direction for future exploration.}

\begin{table}[]
\centering
\setlength{\tabcolsep}{3pt}
\small
\begin{tabular}{@{}R{2.6cm}C{1cm}C{1.2cm}C{0.9cm}C{0.9cm}@{}}
    \toprule
     & {GPT-4o} & {GPT-3.5} & {Llama} & {Claude} \\ \midrule
    {\ourmethod} & 925.34 & 448.77 & 296.28 & 890.37 \\
    {\llmllm} & 929.33 & 414.13 & 290.71 & 706.50 \\
    {\llmsimple} & 914.05 & 418.15 & 298.40 & 601.77 \\ \bottomrule
    \end{tabular}%
    \caption{Average length of summaries, in words, from \ourmethod, \llmllm and \llmsimple with various LLM. For GPT-4o, GPT-3.5, and Llama, we observe no significant difference across methods.}
    \label{tab:lengths}
\end{table}

\subsection{Evaluation of Coverage of Relevant Source Material in \emph{Query-Focused} Multi-Document Summarization}
\label{sec:relevance-relevance} 

\paragraph{Adapting DPPs to select relevant content to a user intent leads to better relevant coverage too} 

\begin{figure*}[ht]
    \centering
    \includegraphics[width=\linewidth]{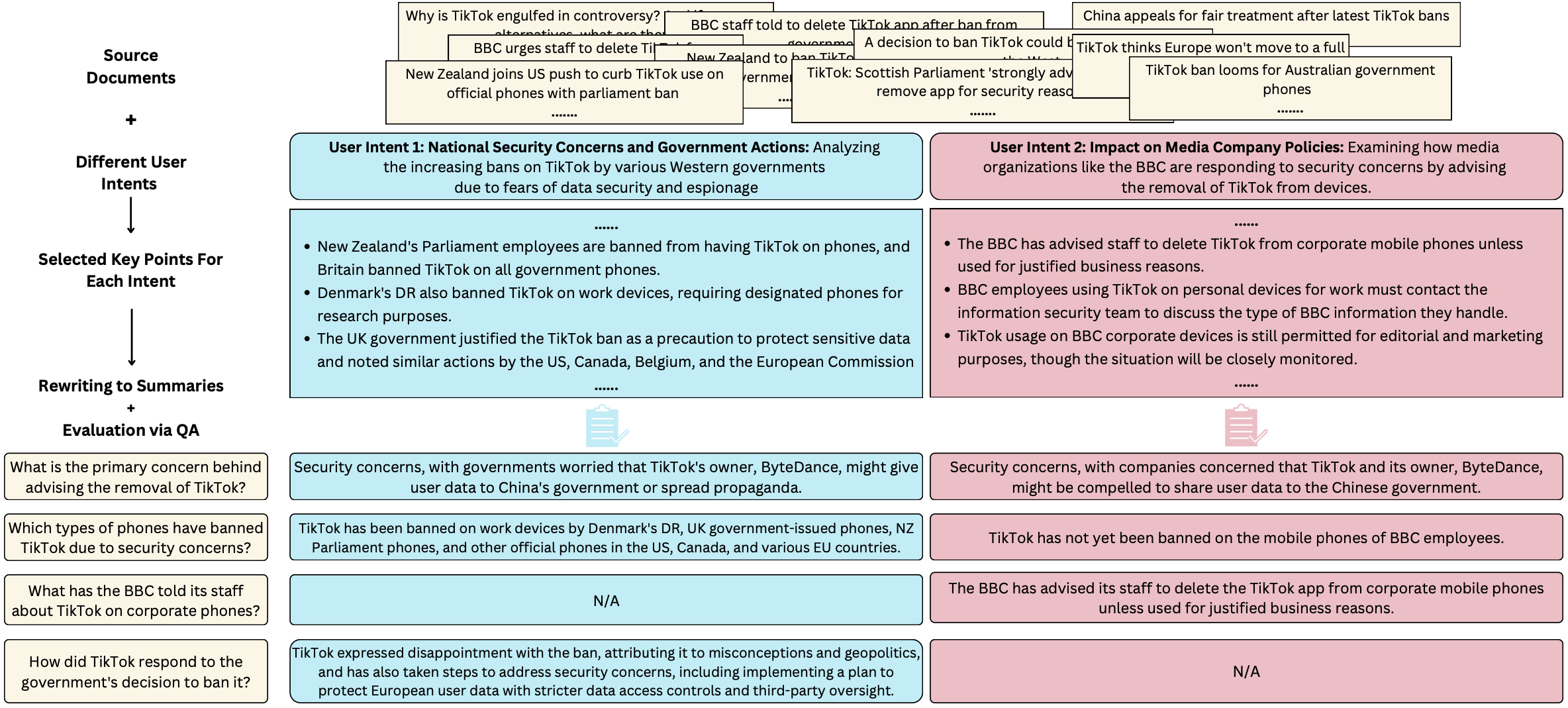}
    \caption{Case study of \ourmethod (\Cref{sec:relevance-relevance}) selecting key points that are diverse and yet relevant to two different user intents (\Cref{sec:kp_rel_selection}) and evaluation of the summaries via question-answering (\Cref{sec:evaluation_method}).}
    \label{fig:example_relevance}
\end{figure*}

\begin{figure*}[ht]
    \centering
    \includegraphics[width=\linewidth]{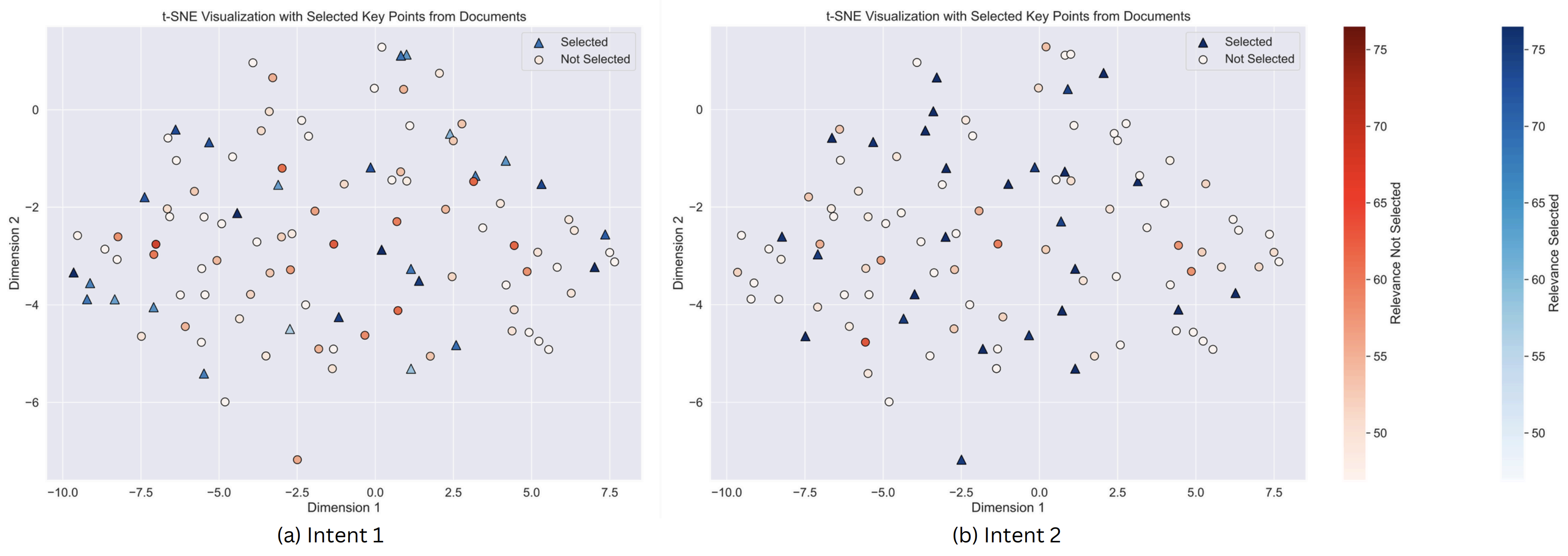}
    \caption{TSNE visualization of the key points selected for the two user intents in \Cref{fig:example_relevance} from the document set. Blue triangles represent selected key points, while red circles denote unselected points. Color intensity reflects relevance to the respective user intent. \ourmethod is able to select relevant key points while also prioritizing diverse coverage of the source material.} 
    \label{fig:relevance_tsne}
\end{figure*}

In addition to ensuring better coverage of the source material, we also evaluate the effectiveness of our proposed method in covering content relevant to specific user intents using the \diversumm Relevance dataset (\Cref{sec:diversumm_intent}). This problem requires balancing diversity of content selected along with relevance to user intent.
From \Cref{tab:results_relevance}, we find that adapting the DPP kernel to incorporate relevance (\Cref{sec:kp_rel_selection}) leads to the highest performance compared to the various baselines. While prompting an LLM to directly generate summaries tailored to user intents (\llmllm) yields improved relevance coverage compared to the naive summarization baseline (\llmsimple), our approach, which combines principled key point selection with relevance-aware DPPs, consistently outperforms both baselines.\footnote{We use prompts \Cref{sec:prompt_for_llmsimple_relevance} and \Cref{sec:prompt_for_llmllm_relevance} for \llmsimple and \llmllm in \Cref{sec:relevance-relevance}.}

\begin{table}[ht]
\centering
\small
\begin{tabular}{@{}R{2cm}cccc@{}}
\toprule
      & \multicolumn{4}{c}{\diversumm Relevance}  \\\midrule
      & GPT 3.5& GPT 4o& Claude& Llama     \\
      \midrule
\llmsimple     & 0.4080 & 0.6410& 0.5843& 0.3182  \\
\llmllm & 0.5292 & 0.6443& 0.6180& 0.3603     \\ 
\midrule
\ourmethod     & 0.5229 & 0.6605& 0.6672& 0.4224 \\
\ourmethod - Relevance & 0.5409 & 0.6972& 0.6937& 0.4501   \\
\bottomrule
\end{tabular}
\caption{Evaluation of coverage of \emph{relevant} source material on \diversumm Relevance (\Cref{sec:diversumm_intent}). We compare the performance of \ourmethod with two relevant baselines (\Cref{sec:models}) across various LLMs. Incorporating relevance into the DPP-kernel (\Cref{sec:kp_rel_selection}) results in the highest coverage, improving over \llmllm prompted to select relevant key points and \ourmethod prioritizing diversity alone.}
\label{tab:results_relevance}
\end{table}

To further illustrate the effectiveness of selecting diverse yet relevant key points, we provide a qualitative case study. \Cref{fig:example_relevance} is an example of two distinct user intents associated with the same set of source documents, along with corresponding representative key points selected by \ourmethod. As a result, the answers to evaluation questions (\Cref{sec:evaluation_method}) differ based on the summaries rewritten from these selected key points. In \Cref{fig:relevance_tsne}, we present a t-SNE visualization of key points from the source documents, embedded using the \texttt{intfloat/e5-mistral-7b-instruct} model (\Cref{sec:models}), that also highlights their relevance to the two user intents and marks those selected by \ourmethod. We observe that \ourmethod effectively balances diverse coverage across the latent space while maintaining high relevance to user queries.

%% file: related.tex
\section{Background and Related Work}
\label{sec:related}
\subsection{Multi-Document Summarization}
\label{sec:related-multi-doc-summ}
Our work builds on foundational multi-document summarization methods that extract information at various granularities \cite{radev2004centroid, hong2014improving, cheng-lapata-2016-neural} and abstractively summarize documents with specialized neural networks \cite{mckeown1995generating, radev1998generating, barzilay1999information, zhang2018towards, fabbri2019multi, song-etal-2022-improving}. This has been aided by various datasets \cite{over2004introduction, dang2005overview, owczarzak2011overview, fabbri2019multi, lu-etal-2020-multi-xscience}, most recent of which is \diversumm \cite{huang2024embrace}. %
More recently, \citet{bhaskar2023prompted, chang2024booookscore} prompt LLMs to hierarchically generate summaries. \citet{to2024skt5scisumm} generate an extractive summary using K-means clustering of sentence embeddings and then rewrite it as an abstractive summary using a fine-tuned T5 model. 
With LLMs able to process longer contexts, \citet{huang2024embrace} primarily evaluate a version of the \llmsimple baseline reporting results on various models. Our work extends this line of research by integrating a prompting pipeline with a principled content selection mechanism using Determinantal Point Processes (DPPs). This approach allows us to combine the strong off-the-shelf generative capabilities of LLMs on the extraction and rewriting subtasks with a robust content selection strategy.  %

\subsection{DPPs for Summarization}
\label{sec:related-dpps-summarization}
Earlier works that use DPPs for summarization tend to be extractive in nature. \citet{kulesza2012determinantal} propose a method to use DPPs for selection of sentences to construct a summary that best resembles the reference in training data, computing the similarity kernel between sentences via TF-IDF scores. \citet{cho-etal-2019-multi} propose to use DPPs to select sentences to construct an extractive summary based on a BERT-based similarity measure. \citet{cho-etal-2019-improving} propose an enhanced similarity metric to further refine extractive summaries. Moving beyond sentence-level extraction, \citet{perez2021multi} introduced DPPs into the attention mechanisms of LSTMs and transformers for abstractive summarization, encouraging diversity in attending to input tokens during generation.
Our method requires no additional fine-tuning, as we make no changes to the model architecture or objective function, unlike previous abstractive methods, allowing us to reap the benefits from further advancements in language modeling. Unlike existing extractive methods, which focus on selecting context-dependent sentences from the documents, we operate on context-independent key points to ensure more high-quality content selection. 

\subsection{Further applications of DPPs}
\label{sec:related-dpp-apps}
DPPs are used in recommender systems when diversity in retrieved items is desirable \cite{lu2020enhancing, wilhelm2018practical}. 
DPPs are also used to select diverse and high-quality in-context learning examples leading to improved performance when prompting LLMs  \cite{wang2024effective, ye2023compositional, yang2023representative}. Finally, DPPs have also been used to help search the prompt space, thereby eliciting jailbreaks of LLMs \cite{zhang2024dpp}. 

%% file: conclusion.tex
\section{Conclusion}
\label{sec:conclusion}
In this work, we demonstrate the utility of explicit content selection for improving the coverage of diverse sources on the \diversumm benchmark. Creating a pipeline that uses LLM prompting steps, for extracting and rewriting information, combined with principled key point selection with DPPs yields summaries that cover diverse source material as well as can be personalized to different user intents. As agentic workflows are increasingly deployed for complex tasks, our findings highlight the need to identify and incorporate principled techniques and tools as a complement to powerful LLMs in order to best suit user needs.  

\section*{Limitations}
\label{sec:limitations}

Firstly, we note the limitations of automatically evaluating coverage on \diversumm with an LLM. While ultimately the gold standard, conducting human evaluations for all ablations is prohibitively expensive, particularly as our task would require annotators to review entire news articles. We followed the evaluation recommendations from \citet{huang2024embrace} and supplemented our automatic evaluation with human validation of the metrics in \Cref{sec:human_validation_augmentations}. Another limitation of this project is that we run experiments on only one dataset, with synthetic augmentations. The main reason for this is that we are intentionally looking for datasets that involve long documents with diverse source material. The challenge with many other summarization datasets is that LLMs already obtain fairly high performance when compared against the references \cite{goyal2022news}. It is yet unclear if our findings would generalize beyond the news domain, and to other languages. We do not make an exhaustive comparison with all possible prompting pipelines for multi-document summarization. Our research question in this project is about evaluating the role of principled content selection in improving coverage so we compare to baselines that do this implicitly (\llmsimple) or via an LLM prompt (\llmllm). It is unclear if this is the maximum performance that can be obtained on the task with a multi-step LLM pipeline. One potential risk from our pipeline is that in \Cref{sec:kp_div_selection}, we select key points purely based on diversity---we do not incorporate any information about the reliability of the particular news articles. Since our work is purely academic, with publicly available datasets, this is not as much an issue but incorporating reliability into systems is important if deployed with real users. 

%% file: appendix.tex
\section{Prompts Used}
\label{sec:prompts}

All prompting experiments were done by sampling from the LLM with temperature $0.7$. We run inference on Claude-3-Sonnet, GPT-4o and GPT3.5 via their APIs and Llama 3.1 with model parallelism on three A100 GPUs.

\subsection{Augmenting \diversumm with synthetic questions}
\label{sec:prompt_additional_questions}
\begin{verbatim}
Write down 10 factual questions that can
be answered from the article below. These 
questions, and their answer should relate
the most important facts of the event 
being reported in the article. Include 
questions that require reasoning about 
the facts in the document. 
Make sure you create questions
such that all the important information 
in the document appears in the answers. 
Each question should be up to 14 words. 
Return a numbered list of questions 
with answers and nothing else.
Article:
<ARTICLE>
    
\end{verbatim}

\subsection{Generating key points from articles}
\label{sec:prompt_extract_kp}
\begin{verbatim}
Summarize all the content in this article
into a list of simple, one-sentence,
bullet points. Make sure that each bullet
point is atomic and can be understood
without any external context. Also, make
sure that all the information in the
article is covered in the list.
Article:
<ARTICLE>    
\end{verbatim}

\subsection{Rewriting the set of selected key points into a coherent summary}
\label{sec:prompt_rewriting}
\begin{verbatim}
Read the following set of key points
obtained from a set of news stories about
a specific topic. From the set, you have
a subset of selected key points. Rewrite
the selected key points into a coherent
report that includes all the details
present in the key points. Make sure the
summary is fluent and coherent. Elaborate
when you summarize diverse or conflicting
information. Make sure to include all of
the factual details from the key points
because we want to use the report to
answer questions. Remember, your output
should be a summary that discusses and
elaborates on the diverse and conflicting
information presented across the articles.
You need to elaborate on the differences
rather than only mentioning which topic
they differ. Don't worry about the summary
being too lengthy. You must give your
response in a structured format:
```Report: [your report]```, where 
[your report] is your generated report.
--------
SELECTED KEY POINTS
--------
<SELECTED KEYPOINTS>
\end{verbatim}

\subsection{\llmsimple baseline prompt}
\label{sec:prompt_for_llmsimple}
We largely reuse the prompt as provided by \citet{huang2024embrace}.
\begin{verbatim}
Read the following news articles. Produce 
a summary that only covers diverse
and conflicting information across the
following articles, without discussing 
the information all articles agree upon.
Elaborate when you summarize diverse or
conflicting information by stating what
information different sources cover and
how is the information diverse or
conflicting. You must give your answer in a
structured format: ```Report: 
[your report]```, where [your report] is
your generated report.
---------
ARTICLES
<ARTICLES>
---------
Remember, your output should be a summary
that discusses and elaborates on the
diverse and conflicting information
presented across the articles. You need
to elaborate on the differences rather 
than only mentioning which topic they  
differ. Don't worry about the summary
being too lengthy.     
\end{verbatim}

\subsection{\llmsimple baseline prompt with relevance}
\label{sec:prompt_for_llmsimple_relevance}
\begin{verbatim}
Read the following news articles and 
associated user intent. Produce 
a summary that only covers the diverse
and conflicting information across the
following articles relevant to the user
intent, without discussing 
the information all articles agree upon.
Elaborate when you summarize diverse or
conflicting information by stating what
information different sources cover and
how is the information diverse or
conflicting. Balance diversity of content 
with relevance to user intent. You 
must give your answer in a
structured format: ```Report: 
[your report]```, where [your report] 
is your generated report.
---------
ARTICLES
<ARTICLES>
---------
USER INTENT
<USER INTENT>
---------
Remember, your output should be a summary
that is relevant to the user intent and
discusses and elaborates on the
diverse and conflicting information
presented across the articles. You need
to elaborate on the differences rather 
than only mentioning which topic they  
differ. Don't worry about the summary
being too lengthy.     
\end{verbatim}

\subsection{Evaluation of source coverage}
\label{sec:prompt_eval_answerability}
\begin{verbatim}
Please act as an impartial judge and
evaluate the quality of the response
provided by an AI assistant. Your
evaluation should consider coverage of
the summary with regard to the question
and answers (i.e. how much information
in the question and answers is covered
by the summary). Begin your evaluation
by deciding if the question is
answerable from the summary - this 
should be a true or false answer. Be as
objective as possible. You next need to
evaluate if the information to answer a
question from the summary matches the
reference answer. The answer to whether
the answer matches should be “0” for
insufficient coverage, and 1 indicates
sufficient coverage. The output should
strictly be in the format of a JSON with
two keys, 'answerable' with the value
true or false, and 'coverage' with the
answer 0 or 1. Return nothing else.
--------
Model Generated Response:
<SUMMARY>
--------
Question:
<QUESTION>
--------
Reference Answer:
<REFERENCE ANSWER>
\end{verbatim}

\subsection{\llmllm baseline prompt}
\label{sec:prompt_for_llmllm}
\begin{verbatim}
Read the following set of key points
obtained from a set of news stories about
a specific topic. From the set, you have
a select a subset that ensure maximum
coverage of the articles provided. 
Make sure that all the important factual
details from the articles are covered
in the selected key points. Ensure that
you  cover all of the diverse viewpoints
mentioned in the articles. Your output
should be a list of selected key points
where each selected one identically 
matches the corresponding key point 
You must give your
response in a structured format:
```Selected Key Points: [your list]```.
--------
KEY POINTS
<ALL KEYPOINTS>
---------
ARTICLES
<ARTICLES>
---------

\end{verbatim}

\subsection{\llmllm baseline prompt with relevance}
\label{sec:prompt_for_llmllm_relevance}
\begin{verbatim}
Read the following set of key points
obtained from a set of news stories about
a specific topic and the associated user
intent. From the set, you have
a select a subset that are relevant to the
user intent and ensure maximum
coverage of the articles provided. 
Make sure that all the important factual
details from the articles that are 
relevant to the user intent are covered
in the selected key points. Ensure that
you  cover all of the diverse viewpoints
mentioned in the articles. Your output
should be a list of selected key points
where each selected one identically 
matches the corresponding key point 
You must give your
response in a structured format:
```Selected Key Points: [your list]```.
--------
KEY POINTS
<ALL KEYPOINTS>
---------
ARTICLES
<ARTICLES>
---------
USER INTENT
<USER INTENT>
---------
\end{verbatim}

\section{Validation of \diversumm Augmented and \diversumm Relevance}
\label{sec:human_validation_augmentations}
In order to confirm that our synthetic augmentations of \diversumm are valid, we perform an additional human annotation. The annotators for this task were volunteer PhD students recruited from our university in the US.

\subsection{Confirming that the questions generated in \Cref{sec:diversumm_aug} are valid}
\label{sec:diversumm_aug_validation} 
We randomly sample 100 LLM-generated question-answer pairs and the corresponding articles from which they were generated. Two separate human annotators independently provide a binary annotation that the question can indeed be answered by the article in question. Both annotators agree that the generated questions are answerable in 98\% of cases. They are also asked to score if the provided answer correctly answers the question given the article. Agreement on the correctness of the provided answer is 93\%.

\subsection{Filtering of synthetic user intents generated in \Cref{sec:diversumm_intent}}
\label{sec:diversumm_intent_filtering}
To create the query-focused version of the dataset, we prompt the model to generate $5$ distinct user intents. For each of the intents, we identify the set of relevant questions by scoring the relevance of all \diversumm-Augmented questions to that particular intent with the trained \texttt{intfloat/e5-mistral-7b-instruct} model. We select this model due to its strong performance on the MTEB leaderboard. We set the threshold as $0.6$ above which a question is deemed relevant. We retain all user intents that contain at least $20$ different relevant questions associated with them. As a result, the average number of user intents evaluated per example is $4.65$ with a minimum of $2$ and a mode of $5$. 

\paragraph{Confirming the validity of synthetic user intents generated in \Cref{sec:diversumm_intent}}
\label{sec:diversumm_intent_validation}
We randomly sample 50 examples, and the associated user intents, out of those that maintain $5$ intents after filtering (\Cref{sec:diversumm_intent_filtering}). These are independently annotated by two separate human annotators. Each annotator provides a score from 1-5 to assess that each individual intent is valid given the set of input documents. The mean rating assigned to the generated user intents is $4.35$ out of $5$, with a Cohen's Kappa of $0.64$ indicating moderate to high agreement. This value also corresponds with the score assigned for \emph{Applicability} of LLM-generated user personas in \cite{balepur2024boat}.  
We then ask the annotators to score the effective number of distinct personas out of the provided $5$, an integer value from $1$ to $5$. Annotators report an average value of $3.56$ indicating that further exploration is necessary in order to synthetically create diverse user intents. 

\section{Additional Results}
\label{sec:additional_results}

\subsection{The latent representations also contain useful information over selecting key points with uniform random sampling} 
\begin{figure}
    \centering
    \includegraphics[width=\linewidth]{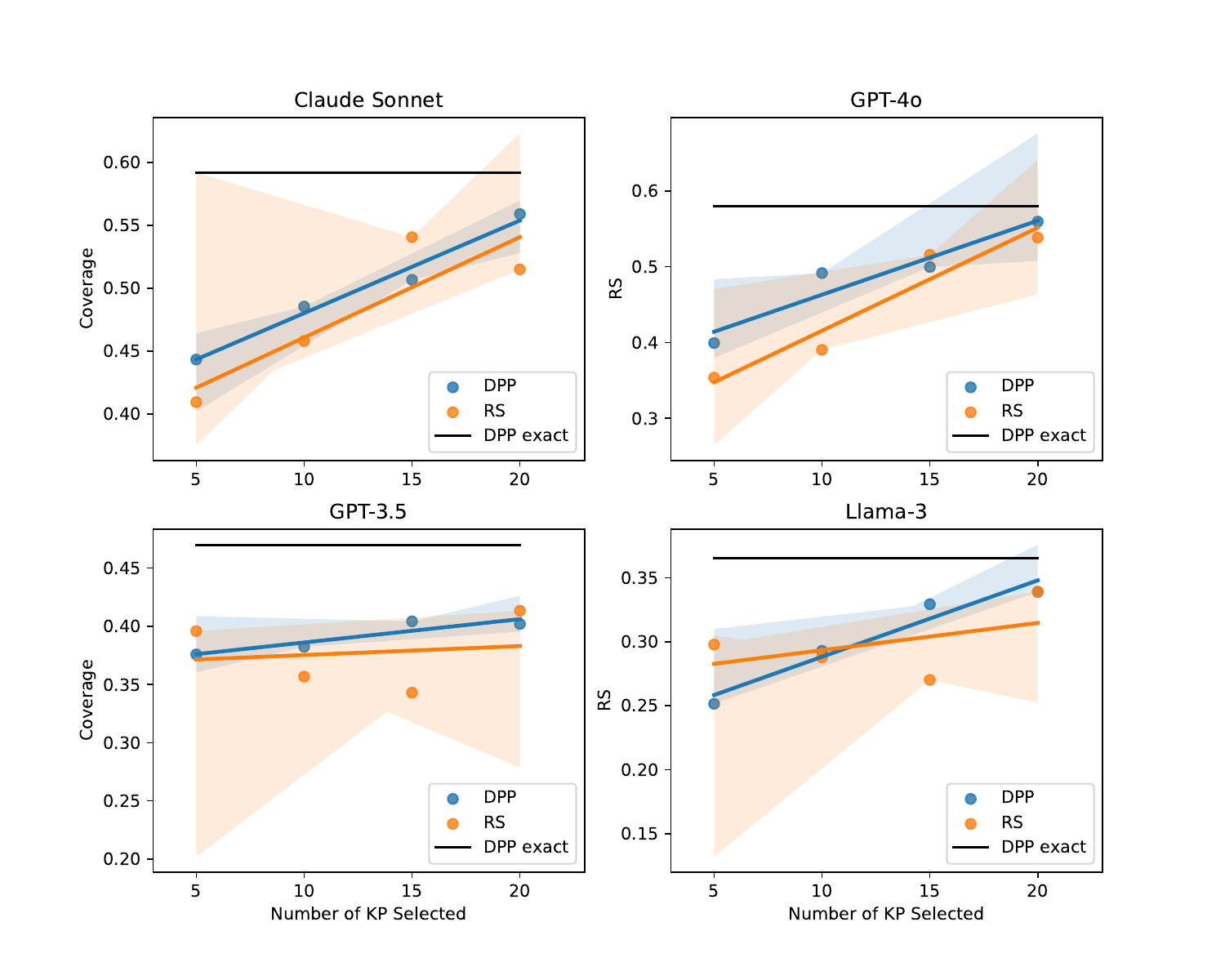}
    \caption{Comparing using a $k$-DPP with uniform random sampling for key point selection for \diversumm, varying $k$, across 4 different LLMs (\Cref{sec:additional_results}). The $k$-DPP consistently outperforms uniform random sampling, showing the value in sampling while considering the learned representations of key points. We also note that both are outperformed by \ourmethod with exact sampling.}
    \label{fig:dpp_vs_urs}
\end{figure}

From \Cref{sec:multi-doc-summ-results}, we observe that prioritizing diversity when selecting key points leads to high coverage in summaries, more uniformly covering all the different source documents. However, uniform random sampling is another way in which we can, in theory, cover each source document. We concatenate key points from all the documents and then randomly sample $k$ of them, before rewriting these into the summary using the prompt in \Cref{sec:prompt_rewriting}. We then compare this baseline with one that selects $k$ key points using a $k$-DPP to represent these. From \Cref{fig:dpp_vs_urs}, we see that, for the same number of key points, the $k$-DPP baseline fairly consistently outperforms uniform random sampling. This again highlights the value in using the learned representations to select key points as it allows our method to sample prioritizing the relative similarities of different key points. Finally, we note that both these methods are comfortably outperformed by the \ourmethod baseline, essentially a DPP with exact sampling as detailed in \Cref{sec:kp_div_selection}. The main difference is that exact sampling sets the number of key points to be selected by considering the nature of the latent space of the key points, and not as a hyperparameter input to the method. This confirms the benefit of combining LLM-prompting with principled techniques as appropriate to achieve high performance on tasks such as \diversumm.

\subsection{Content selection with DPPs on summarized bullet points outperforms extracted sentences from the documents} We also evaluate using a DPP-based extractive baseline that selects extracted sentences from the various articles to prioritize content diversity \citep{cho-etal-2019-multi}. On DiversSumm, coverage with GPT-4o is $0.5090$, vs. $0.5805$ for selecting key points, and with Llama, it is $0.2926$, vs. $0.3653$ for selecting key points. The primary reason for this difference is that key points can synthesize information from different sentences, making them more reliable automatic units of information for selection, a similar finding to \citet{zhang2023extractive}. This finding again highlights the benefit of combining principled tools like DPPs with LLMs in pipelines, leveraging the strengths of both kinds of methods in sequence to obtain high performance. 

\subsection{Tests for statistical significance}
\label{sec:significance-tests}
To evaluate the significance of the coverage improvements shown in \Cref{tab:results_cov}, we perform a two-tailed $t$-test comparing the mean coverage of \ourmethod to \llmllm, the highest-performing baseline, for \diversumm and \diversumm-Augmented. We find that \ourmethod achieves significantly higher coverage than \llmllm for GPT-3.5, Claude, and Llama 3.1 at the 5\% significance level ($p < 0.05$). For \diversumm-Augmented, the improvement is significant for all four LLMs, likely due to increased statistical power from the larger sample size. Similarly, for \Cref{tab:results_relevance}, we perform a two-tailed $t$-test comparing \ourmethod with and without relevance in the DPP-kernel. Incorporating relevance leads to significantly higher coverage ($p < 0.05$) across all LLMs.